# AUTOMATIC KNOWLEDGE ACQUISITION
# FOR OBJECT ORIENTED EXPERT SYSTEMS

Joël COLLOC, Danielle BOULANGER

URA CNRS 1257
IAE, Université LYON III

15, Quai Claude Bernard
BP 0638 69239 LYON CEDEX 02
TEL : 72.72.21.58   FAX : 72.72.20.81

**ABSTRACT**

We describe an Object Oriented Model for building Expert Systems.
This model and the detection of similarities allow to implement reasoning modes as induction, deduction and simulation.
We specially focus on similarity and its use in induction.
We propose original algorithms which deal with total and partial structural similitude of objects to facilitate knowledge acquisition.

**RESUME**

Nous décrivons un modèle O.O pour construire des systèmes experts.
Ce modèle et la détection de similarités permettent d'implanter des modes de raisonnement comme l'induction, la déduction et la simulation.
Ce travail concerne surtout la similarité et son utilisation dans l'induction.
Nous proposons des algorithmes originaux capables de détecter une similarité de structure totale ou partielle entre des objets pour faciliter l'acquisition de connaissances.

**Mots clés** : Acquisition de connaissances, modèle orienté objet, similarité structurale, induction.

# 1. INTRODUCTION

This paper proposes an object oriented model for building expert systems. While this model enhances the knowledge modularity, it supports some other reasoning modes than traditional deduction.
First, we present the characteristics of our object oriented model (COLL 89), then we highlight the features used to implement reasoning and allow knowledge acquisition.
Next, we show how to deal with similarity between complex objects and with induction of new properties (object structures, attributes and functions).
The induction dynamically enriches object types in an incremental way. Then, new type properties are available for being inherited by old and new instances.
Evaluation functions, integrating comparison algorithms provide means to find structural similarities in objects.
Similarity should be also detected by comparing object properties (attributes and methods) (REIT 87)(GALL 88). However, this feature will be described in future researches.
Similarity is the key concept for both deduction and induction reasoning. Simulation uses dynamic functions (with time parameters) and is briefly presented in this work.

# 2. WORK HYPOTHESIS

We describe an object oriented model which is used to represent knowledge and some fuzzy reasoning modes like induction, deduction and simulation, all exploiting similarity concept.
Although rule systems are often appropriate in deduction implementation, they provide few features to deal with other reasoning modes (ADAM 89)(HARM 90)(BART 90). The model employs encapsulated evaluation functions which search for structural similarity (sub-objects) and matching attributes.Evaluation functions compute instance conformity with structural domains, integrity constraints supplied by object types. The O.O. model supports an O.O. method for expert system design (BOUL 93). This method guides the knowledge engineer and the experts to represent their perception of knowledge domains in term of objects. The object modularity provides means to integrate several O.O. knowledge bases.
In this paper, we only focus on similarity concept and induction.

## 2.1. Similarity concept

Similarity is involved in every reasoning modes used by human beings because it allows to recognize all objects in our environment. We continually compare the perception of world object properties with the abstraction we have built from them, all life long. We naturally know what properties are necessary to recognize an object and to classify it in the proper abstract class identified by a name. When an expert must find similarities between abstract objects which have no other existence than in his own mind, he compares the properties (structure, behaviour) of these objects, in a very similar way. The difference lies in the fact that abstract object properties are not perceived but built up in mind and then memorized. Similarity is a key concept because it is used by both deductive and inductive reasoning. (TSOT 87)

## 2.2. Inductive reasoning

Inductive reasoning allows us to try to extend the specific observations concerning a unique object of the world to the whole object class it belongs to.
Thus, induction utilizes generalization because knowledge obtained from one instance has a strong probability to be observed in all other objects of the same type. The use of induction is especially evident in the scientific fields when the occurrence of singular cases often give opportunities to increase our knowledge by finding new behaviour and features for a whole set of similar instances.

The new observed properties can be either linked to the set of the class properties (because there is a sufficient similarity found out between the unique case and the others) or attached to a new derived class whose unique case will be the first instance.

Researchers use induction to build theories by setting then proving hypothesis and at last generalizing the results to a set of similar cases.

## 3. AN OBJECT ORIENTED MODEL (figure 1)

In our model (COLL 89), the objects are nested in each other. According to the applications, a unit corresponding to a reference object type (or level 0) has to be defined : each instance of this type is called unique object.
This notion defines a boundary between the internal level and the external level : all the component objects belong to the internal environment, inside the "unique object"; while the others reside in the external environment. Objects that contribute to build the unique object, are called sub-objects.
The internal structure represents the content of the object and its composing sub-objects which determine a partition of set of properties.
The external environment expresses the relationships of an object with the others.
Each sub-object can be seen, at any time, as a unique object. Its proper sub-objects build up its internal structure. We call zoom effect this adapted perception of objects.
The model provides a twin conceptual level. (figure 1)

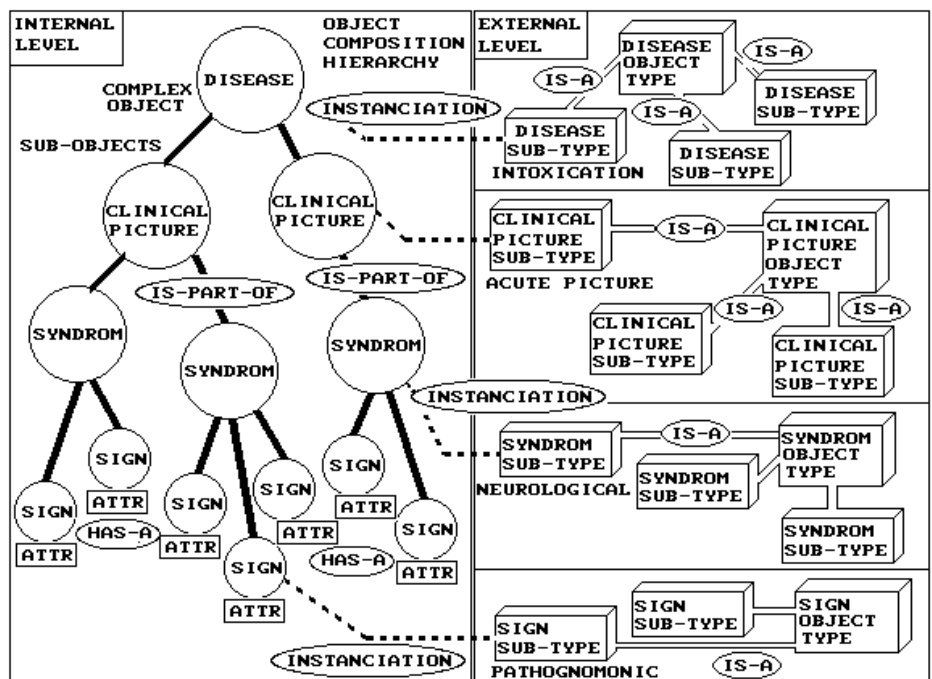
Figure 1 :Model, the twin conceptual level example : the disease concept description

3.1. The internal level or "inner object"

It includes :
- aggregation relationships of sub-objects (IS-PART-OF)
- relationships grouping all the attributes of an object (HAS-A).
Aggregation relationships establish a multiple upward inheritance which passes composing sub-objects properties on to "compound object".
The internal level encapsulates attributes and functions which represent the static characteristics and the behaviour of the objects.
The constraints on the objects are either static (constraints on the structures, the attributes, the cardinalities) or dynamic; in this case, they are functions of time.
The spontaneous evolution of the object state is translated by dynamical functions of the inner object. These functions automatically modify the composition links and the reference values of the attributes, at any time.
Evaluation functions of the "inner object" compare the structure of an object (on its composing sub-objects), the value of particular attributes to these of the other objects.

## 3.2. The external level or outer-object

It includes the generalization / specialization relationships (IS-A) between object types and object sub-types. These relationships establish a type hierarchy and a simple descending inheritance which passes object type properties on to its sub-types.
The global evaluation functions belonging to instances of the application object type, assess all the object type instances involved in an application.
The model makes a distinction between type concept (object structure and properties) and class concept (a set of instances).

## 3.3. Links between external level and internal level

### 3.3.1. Instantiation

The internal and external levels are linked by the instantiation mechanism (figure 1) which determines new objects in accordance with each type.
The qualifier attributes are particular attributes (listed at the external level as characteristics of the object type) whose instantiation is mandatory at the internal level. All attributes of an object are not "qualifier attributes". In this case, the value attribution is optional.
Communication between objects is performed by messages and interface functions. Messages carry information from an emitter object to one or several receiver objects.

### 3.3.2. Encapsulation and interface functions

The input (or affectation) functions verify the values to be given to the attributes of the inner object and then affect them.
The output (or reading) functions give us the state of an object. They are activated by the reception of a valid input-message. Then, they reach the value of some attributes and finally provide an output message carrying attribute values. The owner of the object chooses the attributes which can be read from the external environment; the others will stay in the internal environment as private attributes.

### 3.3.3. Messages features

When objects need to communicate and when they are not connected by any composition relationship, they must use messages. Messages are sent by a transmiter object toward one or several receiver objects named targets. Messages will be especially used by application objects to gain information from knowledge objects.

### 3.3.4. Application object type

The application object type is a predefined system type, whose aim is to describe applications which suit with several user needs.
Objects of application type encapsulate :
. data - metaattributes are used to represent intermediate states and results of a treatment.
. metafunctions are designed to define existing relationships, else than composition, which exists between domain objects of some types, involved in an application.
They belong to the internal environment of application objects.
Metafunctions send messages to obtain the internal state of these objects and store the received result in metaattributes.

The original points of this model are :
- distinction between the internal environment and the external environment for object description.
- multiple ascending inner inheritance without conflicts because it exists a partition on the sub-object set at each level of the composition hierarchy. Generally this feature is not provided by O-O models (except NEUH 88).

- evaluation and dynamical functions which provide means to assess object state and to model object evolution. Evaluation functions make use of partial similitude algorithms.
- application object type

*NB : the methods encapsulated in the objects are called functions in our model*.

This O-O model offers semantic capability, the application object type concept allows to implement similarity algorithms which provide semantic relativism.

## 4. IMPLEMENTING SIMILARITY WITH THE O.O. MODEL

4.1. Development of similarity for reasoning purpose

The object oriented model, by the means of object types allows to take into account the whole set of objects concerned by a problem resolution in a given domain concurrently. The aim of the model is to represent knowledge in terms of objects (BOUL 88) (COLL 92)

Expert systems using this knowledge deal with the whole object properties of a reference type, not successively but simultaneously in a single reasoning cycle.

Besides the model allows to find common points between objects first by comparing their types and second by using evaluation functions which return values representing the similarity degree of their contents.

The model favors similarity reasoning instead of sequential and dichotomic deductions.

Evaluation functions provide a solution for threshold uncertainty by calculating rates of similarity (FU 83)(COLL 87)(MANA 88).

Global evaluation functions implemented in application objects are often discrete. They compare instance structure and properties of objects belonging to a reference type selected by zoom effect (COLL 89).

4.2. Different kinds of similarity

The whole process involve in turn generalization/specialization, object composition and object instantiation. We distinguish two kinds of similarity providing several levels of successive refinements. (ABOU 81)

In a first step, the comparison process tries to make out global or partial similarity between objects. If successful, the second step searches for a value resemblance between "qualifier attributes" of similar objects.

In a final step behavioural similarity tries to compare object behaviour by reviewing methods which belong to the private object part.

We focus on the first step.

4.3 Object similarity algorithms

Structural similarity has to establish that two objects posses similar sub-objects and so on recursively towards the leaves of the composition hierarchy of the two objects. Global or partial structural similarity between objects takes place by comparing the lists of their sub-object types.

Sometimes, only a partial structure similarity involving several sub-objects can be found. As some parts of objects are more important than others, the partial similarity can be very valuable (for example, the lock is the major sub-object of the door).

If we compare OBJ1 and OBJ2 structures whose respective types are TYPE1 and TYPE2 several cases can occur :

**Case 1 : OBJ1 and OBJ2 belong to the same type.**
OBJ1 and OBJ2 have, by definition, a strong degree of structural relationship. But we have to consider that the structural relationship depends of the specialization level brought by the two object type.

Specialization criteria are in order : structure including sub-objects and qualifier attributes, behaviour represented by methods (procedures and functions in our model), values of attributes and functions.

However, if OBJ1 and OBJ2 type is strongly specialized, the inherent structure constraints will be sufficient to establish a valuable comparison between its instances.

**Case 2 : OBJ1 and OBJ2 belong to the same type but have heterogeneous structures**

The sub-objects that compose the object can be selected among several listed distinct object types giving heterogeneous structures.

Legal sub-object types are listed in the type header. We distinguish two categories of sub-objects - those whose existence is essential to the object structure and those which are optional. The first category allows to build the skeleton and the kernel of the concept represented by the object and the second one offers some variants. This distinction prevents the useless multiplication of object sub-types and the misuse of specialization.

To compare the heterogeneous structures of OBJ1 and OBJ2, we have to build an evaluation function which will be included in an application object.

The evaluation function gets two lists of sub-object types. By comparing the two lists, the evaluation function computes a structural similarity score taking into account the relative importance of sub-objects.

**Case 3 : OBJ1 and OBJ2 belong to different types TYPE1 and TYPE 2**

First, we must check if one of the two types is a super-type of the other one. To carry out this operation, the evaluation function has to consult TYPE1 super-type list to search for TYPE2. If unsuccessful, it examines TYPE2 super-types to try to find out TYPE1. If negative again, it comes to the conclusion that TYPE1 and TYPE2 are independent types.

**Case 3: first variant - TYPE1 is a TYPE2 super-type or conversely**

TYPE1 and TYPE2 are connected by an inclusion dependence and OBJ1 and OBJ2 are both TYPE1 instances (the most general type). Thus, if an evaluation function exists for TYPE1, we can use it to compare OBJ1 and OBJ2.

**Case 3 : second variant : TYPE1 and TYPE2 are independent**

We can assume that OBJ1 and OBJ2 show a poor similarity degree. However, we can imagine that these objects have been built independently at different times, by several users. So, we have to verify if they have a partial similitude structure. We use an evaluation function to examine sub-object types involved in OBJ1 and OBJ2 respective structures.

This evaluation function belongs to every object types and is context free. The function implements a recursive comparison algorithm implemented in C code.

The algorithm compares the inner composition hierarchies of two objects and tries to find as many sub-objects whose type match, as possible.

The trees are examined depth first because :
 - the hierarchies have generally a different depth,
 - sub-objects of same complexity have to be compared.

**Step1**

The composition trees of objects OBJ1 and OBJ2 are compared according to the number of sub-objects, each sub-object owns in its composition hierarchy .

**Step2**

So, the algorithm builds two lists of nodes (one for OBJ1, one for OBJ2) sorted in ascendant order according to the respective number of sub-objects each sub-object owns in its composition hierarchy. (see annex)

Each object node shows the following fields : Object identifier, Object type identifier, Number of sub-objects, Father object identifier, First child sub-object for this object, Next child sub-object for father object.

**Step3**

Then the algorithm compares the lists, it counts the sub-objects belonging to the same type and it builds a match table (figure 2) to compute the maximum degree of structural resemblance.

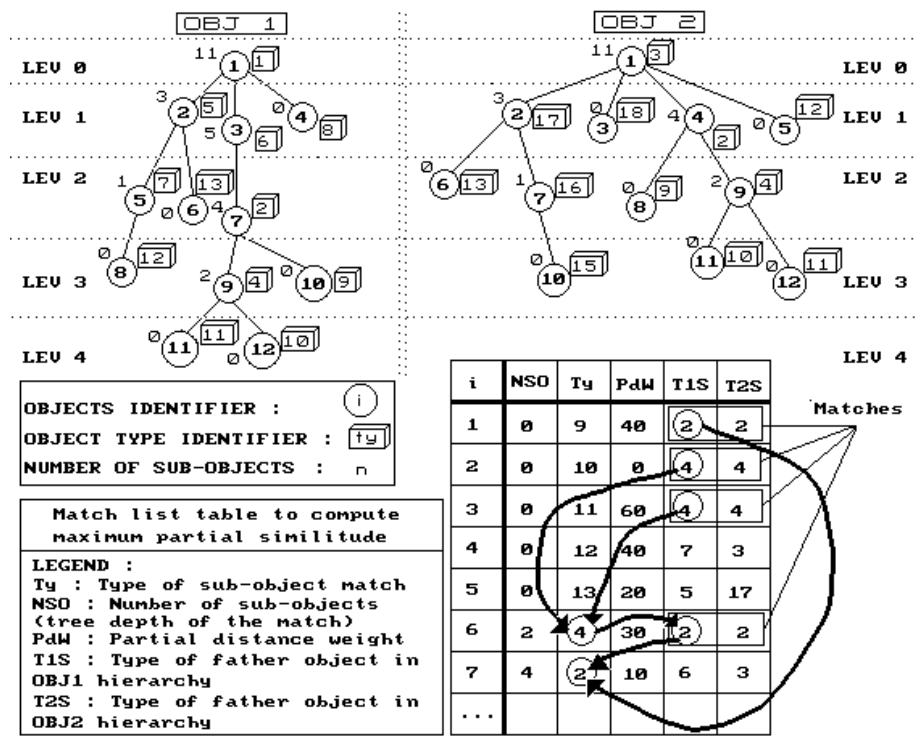

Figure 2 : Comparing the inner hierarchies of two objects

**Step4**
The match table (figure ) shows all sub-object matches sorted according to the depth (NSO).
The algorithm compares the respective types T1S and T2S of father objects of the two matching sub-objects.
If the two sub-objects belong to the same type (that is T1S and T2S show the same identifier) , a new sub-object node of this type is searched in the table by examining the Ty entries. (The weight of the match is computed with the maximum function).
If a match node of this type is found, the process continues until T1S and T2S are found different .
When a dissimilarity has been exhibited the process stops.
**For example :**
Match#2 shows the same father object type identifier T1S=4,T2S=4. Type#4 is searched in Ty column.
Match#6 involves sub-objects of Type#4, and they show the same father object type T1S=2, T2S=2. So Type#2 is searched in Ty column.
Match#7 involves sub-objects of Type#2 but they do not show the same father object type (T1S=6, T2S=3).
The algorithm found the greater similar subtree and so the maximum degree of partial structure similarity. The algorithm repeats the same process for each match node.
The algorithm complexity is $O(n^2 . \log n)$ if $n = \min(n1,n2)$ ; n1 and n2 are the respective number of OBJ1 nodes and OBJ2 nodes.

## 5. INDUCTION WITH THE OBJECT ORIENTED MODEL

5.1. Induction and generalization
Induction operates the generalization / specialization hierarchy provided by the model. Induction is used whether to enrich existing type properties, and possibly their father type ones, or to derive new sub-types taking into account the new induced feature. Induction provides means to implement automatic knowledge acquisition. It makes use of different kinds of similarity, dependent on the property nature we went to induce.
Induction can be applied as soon as a sufficient structural similarity degree exists between two objects. In the algorithm, the degree is computed from partial distance weight parameters provided by the experts at the design of knowledge objects (BOUL 93).

It is a very useful way to find new specialization links that were not evident at first glance. Partial structural similarity between two sub-objects can be used to support induction. So the new induced property is first added to the similar sub-object and to its type, and then, by applying the inner ascending inheritance, to all objects, in the composition hierarchy, including this sub-object.

5.2. Hypothesis (figure 3)
We present an example which shows how to induce a property in an object type by the means of its instance A1 and the similarity found out with another object named B1 of a different type.
Object A1 is instance of TyA type and belongs to the associated class CA={A1,...,A2,...An}. Object B1 is instance of TyB type and belongs to the class CB={B1,...,B2,...Bn} B1 owns the P5 property (for instance a qualifier attribute and its constraints).
We assume that a total or partial similarity has been established between A1 and B1 by the means of the algorithms described above. Induction will try to induce the P5 property of B1 object to A1 object type.

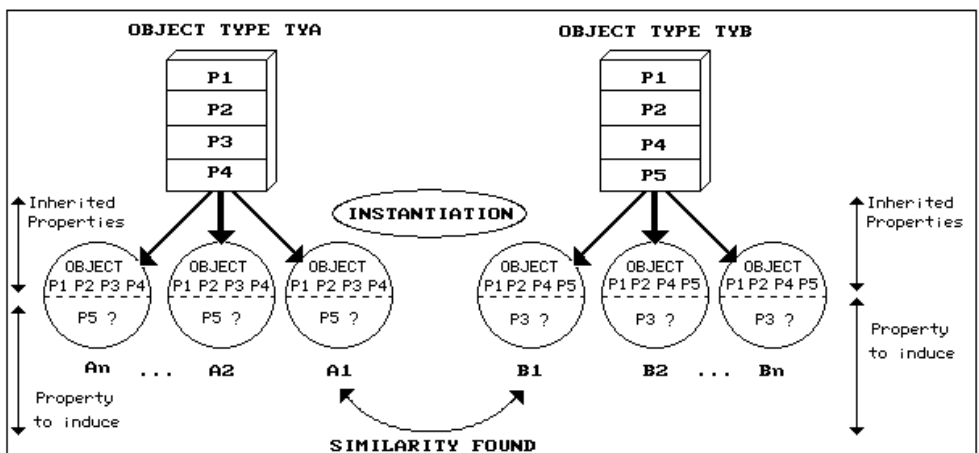
Figure 3 : Induction procedure (initial condition)

5.3. Induction procedure description
**Step 1** : First, the procedure has to specialize TyA type for which A1 is one instance, in the TyA1 type which support the P5 property.
**Step 2** : Then the procedure verifies for each object of TyA, if it should be an instance of TyA1 (figure 4)
**Step 3** : Finally, if all objects of TyA type become instances of TyA1, then TyA is replaced by TyA1 type provided with the new P5 property .
Remark : The TyA type was incomplete and was "able to gain the new P5 property which can be then inherited by all current and future instances.
Else, TyA1 remains as a sub-type of TyA improving the specialization hierarchy.

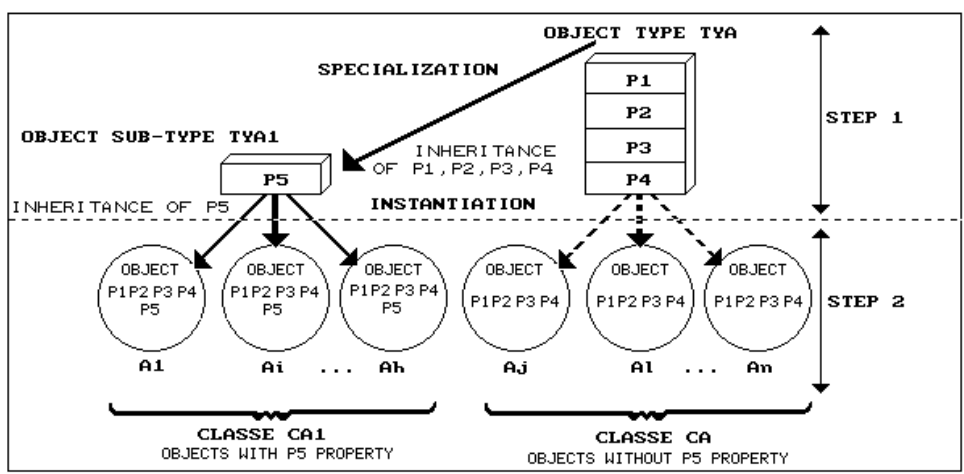

Figure 4 : Induction procedure (step 1 and step 2)

## 6. CONCLUSION AND FUTURE WORKS

Knowledge acquisition is a difficult and time consuming process that often creates a bottleneck in building an expert system.
The work of automating the knowledge acquisition process is in progress, but most tools have been designed for eliciting knowledge from individuals (LIOU 92)

Moreover interaction among experts, creates a synergy that often results in an similar, enriched domain of expertise.
Our propositions are in knowledge elicitation with multiple experts.
Resources and originality of the model have allowed to automate some reasoning modes.
We have especially focused on reasoning modes which make use of similarity and we have provided an original algorithm to detect global and partial structural resemblance of objects.
Induction mechanisms reuse this algorithm and provide means to enrich object types automatically.

Induction shows a new interesting way in learning domain, especially by the means of automatic specialization mechanisms.These mechanisms allow the knowledge base enrichment, the improvment of object type consistence and avoid the creation of useless sub-types.
We hope to extend application fields to other reasoning modes - simulation (tests and backtracking reasoning) with the help of dynamic functions supported by the model.
Deduction is implemented using the similarity concept because it is rarely based on a strict equality of properties.

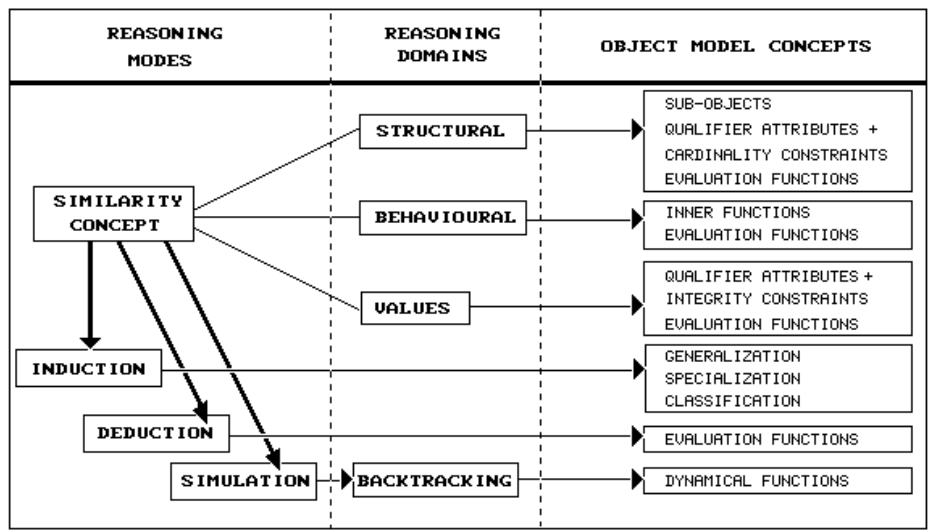

Figure 5 : Reasoning modes summary

We also use similarity algorithms in multidatabase concept to detect heterogeneity provided by pre-existing automated information systems (BOUL 92).

```c
MATCHNODE * compare(LISTE * l1, LISTE * l2, float * wgd, int * nbmatch)
 {  int nbnode = 0;   /* number of nodes examined */
    static MATCHNODE * debut;  /* table pointer */
    BLOC * p1, *p2;        /* list pointers */
    * wgd = 0;         /* global distance initialization */
    debut = calloc(TTABLE, sizeof(MATCHNODE));
    p1= l1->tete;  p2 = l2->tete;    /* p1 pointer for liste l1 , p2 pointer for liste l2 */
 while(p1!=(BLOC *) FIN && p2 !=(BLOC *) FIN)
    {
     while((p2->nso < p1->nso) && p1!=(BLOC *) FIN && p2 !=(BLOC *) FIN)
         { /* search for same kinds of sub-objects */
           * wgd += distance(p1->w, p2->w);  /* compute distance */
           nbnode++;
           p2 = suivant(p2);  }
     while((p1->nso < p2->nso) && p1!=(BLOC *) FIN && p2 !=(BLOC *) FIN)
         { /* search for same kinds of sub-objects */
           * wgd += distance(p1->w, p2->w);  /* compute distance */
           nbnode++;
           p1 = suivant(p1); }
     while((p1->nso == p2->nso) && p1!=(BLOC *) FIN && p2 !=(BLOC *) FIN)
       { /* compare same kinds of sub-objects */
         while((p1->ty < p2->ty) && (p1->nso == p2->nso) && p1!=(BLOC *) FIN && p2 !=(BLOC *) FIN)

          {   /* search for a match 1st case */
           * wgd += distance(p1->w, p2->w);  /* compute distance */
           nbnode++;
           p1 = suivant(p1); }
        while((p2->ty < p1->ty) && (p1->nso == p2->nso) && p1!=(BLOC *) FIN && p2 !=(BLOC *) FIN)
           {   /* search for a match 2nd case */
           * wgd += distance(p1->w, p2->w);  /* compute distance */
           nbnode++;
           p2 = suivant(p2);
           }
        while((p1->ty == p2->ty) && (p1->nso == p2->nso) && p1!=(BLOC *) FIN && p2 !=(BLOC *) FIN)
           { /* match found update the table of matches */
           * wgd += distance(p1->w, p2->w);  /* compute distance */
           nbnode++;  (* nbmatch)++;
           /* dynamically update matches table */
           printf("\n match found %d\n", * nbmatch);
           debut = addtable(debut, nbmatch, p1->ty, p1->nso, distance(p1->w, p2->w),
                     1->perety, p2->perety);
           p1 = suivant(p1);
           p2 = suivant(p2);
           }
      } /* endwhile same nso */
    } /* endwhile END not found */
    while(p2 != (BLOC *) FIN)
       {   /* deal with remaining nodes of l2 list */
         * wgd += distance(p1->w, p2->w);  /* compute distance */ nbnode++;
         p2= suivant(p2);
       }
    while(p1 != (BLOC *) FIN)
       {   /* deal with remaining nodes of l1 list */
         * wgd += distance(p1->w, p2->w);  /* compute distance */ nbnode++;
         p1 = suivant(p1);
       }
  }
```

```c
void partialanalogy(MATCHNODE * p, int * n)
  {
    int i, j ,k, cty, compt;
    printf("\Nombre d'éléments à traiter %d\n",* n + 1);
    for(i = 0; i <= * n; i++)
        {
        j = i;
        compt = 0;
        while( (p+j)->t1s == (p+j)->t2s )
            {
                cty = (p+j)->t1s;   /* current type */
            for(k = 0; k <= * n; k++)
                {
                if (cty == (p+k)->ty);  /* new entry found */ {
                    (p+k)->pdw = cmax((p+k)->pdw, (p+j)->pdw); j = k;
                    compt ++;
                    }
                }
            }
        printf("\n partial analogy was found : two instances");
        printf("\n of %d type number belongs to OBJ1 and OBJ2", (p+j)->ty);
        }
} /* end of procedure */
```